\documentclass{article}
\usepackage{iclr2026_conference,times}

\usepackage{amsmath,amsfonts,bm}

\def\eqref#1{equation~\ref{#1}}

\def\1{\bm{1}}

\DeclareMathAlphabet{\mathsfit}{\encodingdefault}{\sfdefault}{m}{sl}
\SetMathAlphabet{\mathsfit}{bold}{\encodingdefault}{\sfdefault}{bx}{n}

\newcommand{\KL}{D_{\mathrm{KL}}}

\usepackage{hyperref}
\usepackage{url}
\usepackage{booktabs}
\usepackage{amsmath}
\usepackage{amssymb}
\usepackage{mathtools}

\usepackage[utf8]{inputenc}
\usepackage[T1]{fontenc}
\usepackage{amsmath,amssymb}
\usepackage{booktabs}
\usepackage{graphicx}
\usepackage{subcaption}
\usepackage{enumitem}
\usepackage{algorithm}
\usepackage[noend]{algorithmic}

\algsetup{indent=1.1em}

\newcommand{\abar}{\bar{\alpha}}
\newcommand{\Qbar}{\bar{Q}}
\newcommand{\sg}{\operatorname{sg}}
\newcommand{\Cat}{\mathrm{Cat}}

\usepackage[most]{tcolorbox}
\usepackage{xcolor}

\definecolor{examplebg}{HTML}{F7F7F8}
\definecolor{exampleframe}{HTML}{D9D9DE}
\definecolor{corruptblue}{HTML}{3567B7}
\definecolor{corruptbg}{HTML}{EAF1FB}

\newtcolorbox{corruptionbox}[1]{
    enhanced,
    breakable,
    colback=examplebg,
    colframe=exampleframe,
    boxrule=0.5pt,
    arc=2pt,
    left=8pt,
    right=8pt,
    top=7pt,
    bottom=7pt,
    before skip=8pt,
    after skip=8pt,
    fonttitle=\small\bfseries,
    title={#1},
    coltitle=black,
    colbacktitle=examplebg,
    attach boxed title to top left={
        xshift=0pt,
        yshift=-1pt
    },
    boxed title style={
        colback=examplebg,
        colframe=examplebg,
        boxrule=0pt,
        left=0pt,
        right=0pt,
        top=0pt,
        bottom=2pt
    }
}

\usepackage[most]{tcolorbox}
\usepackage{booktabs}
\usepackage{xcolor}

\definecolor{examplebg}{HTML}{F7F7F8}
\definecolor{exampleframe}{HTML}{D9D9DE}
\definecolor{repairbg}{HTML}{EAF1FB}
\definecolor{repairblue}{HTML}{3567B7}
\definecolor{errorbg}{HTML}{FBECEE}
\definecolor{errorred}{HTML}{A94442}

\newtcolorbox{resultbox}[1]{
    enhanced,
    colback=examplebg,
    colframe=exampleframe,
    boxrule=0.5pt,
    arc=2pt,
    left=8pt,
    right=8pt,
    top=7pt,
    bottom=7pt,
    before skip=8pt,
    after skip=8pt,
    fonttitle=\small\bfseries,
    title={#1},
    coltitle=black,
    colbacktitle=examplebg,
    attach boxed title to top left={
        xshift=0pt,
        yshift=-1pt
    },
    boxed title style={
        colback=examplebg,
        colframe=examplebg,
        boxrule=0pt,
        left=0pt,
        right=0pt,
        top=0pt,
        bottom=2pt
    }
}

\title{Learning to Re-Draft: A Variational \\ Stackelberg Game for Discrete Diffusion}  

\author{%
  \textbf{Dmitrii Moor}$^{1}$ \quad
  \textbf{Federico Tomasi}$^{1}$ \quad
  \textbf{Paul N.\ Bennett}$^{2}$ \quad
  \textbf{Alice Wang}$^{3}$ \quad
  \textbf{Mounia Lalmas}$^{1}$ \\[0.5ex]
  $^{1}$Spotify, London, UK \quad
  $^{2}$Spotify, Boston, USA \quad
  $^{3}$Spotify, New York, USA \\
  \texttt{\{dmitriim,federicot,pbennett,alicew,mounial\}@spotify.com}
}

\iclrfinalcopy
\date{\today}

\begin{document}
\maketitle
\fancyhead{}
\renewcommand{\headrulewidth}{0pt}

\begin{abstract} 
Discrete diffusion models offer the ability to re-draft, revisiting and correcting earlier tokens throughout generation.  
This capability depends on the forward corruption process that defines what the denoiser learns to correct. 
Masked diffusion models fix tokens once they are unmasked, while uniform diffusion permits revisions but relies on uniformly random token substitutions. 
We instead learn which substitutions are most useful for training the denoiser to re-draft. We introduce Variational Stackelberg Discrete Diffusion (VSDD), a framework for learning a semantically aware corruption process. 
VSDD formulates training as a leader-follower game: the leader defines a Markovian corruption process parameterized by the denoiser’s token embeddings, while the follower optimizes a variational denoising objective with the corruption process held fixed.  
The leader rewards corruptions based on how much the denoiser improves after learning from them, rather than on how easily the current denoiser can reconstruct them. 
We measure this improvement under a fixed reference corruption process, approximate the follower's response with a one-step gradient update, and optimize the leader using a score-function estimator.
We evaluate VSDD across molecular, text, and playlist generation. VSDD substantially improves molecular validity over uniform and masked diffusion, reduces text perplexity relative to uniform diffusion while remaining competitive with masked diffusion, and achieves sizable improvements in offline playlist recommendation  metrics.

\end{abstract} 
 
\section{Introduction} 
\label{sec:intro}
Discrete diffusion models~\citep{austin2021structured, Hoogeboom2021} have emerged as a powerful paradigm for generating sequences of discrete tokens.
These models learn to reverse a stochastic corruption process that gradually transforms a clean sequence into noise. 
Unlike autoregressive models, which generate tokens sequentially from left to right and can suffer from error accumulation~\citep{BengioVJS15, Ranzato2015SequenceLT}, diffusion models operate over the entire generated sequence and can iteratively revise earlier predictions during denoising.  This ability to \emph{re-draft} is particularly valuable for sequences with strong structural dependencies, such as molecular sequences. 


The performance of discrete diffusion models depends on two closely coupled components: the \emph{denoiser}, 
which predicts the clean sequence from a noisy observation, and the \emph{forward noise process}, which determines how clean tokens are perturbed during training. 
While significant research effort has focused on denoiser architectures~\citep{NEURIPS2024_eb0b13cc, Lou2024, Shi2024}, the forward process is typically fixed to a simple predefined distribution, such as uniform or absorbing noise~\citep{austin2021structured}. 
Yet the choice of the forward corruption process can substantially affect learning efficiency and generation quality. 

For instance, \cite{NEURIPS2024_eb0b13cc} showed that absorbing noise can substantially outperform the uniform diffusion \citep{austin2021structured} for text generation, suggesting that \emph{how} tokens are corrupted can be as important as \emph{how} they are reconstructed.
Similarly, \cite{nichol2021improved} showed that a non-linear noise schedule can further improve generation quality. 
These observations motivate a natural question: can we \emph{learn} a forward noise process that adapts to the data distribution and the denoiser's capabilities?

Learning the forward corruption process, however, introduces a fundamental optimization challenge. 
Recent work takes a step in this direction by jointly optimizing the forward process and the denoiser through a score function based surrogate loss (\citealt{bartosh2026forwardlearned}). 
With such joint optimization, the forward process may improve the training objective by compensating an underperforming denoiser rather than providing corruptions that help the denoiser improve. 
In particular, it may favor trivial corruptions that are easy to reconstruct, reducing the KL-loss without improving the resulting generative model. 
The challenge, however, is to learn corruptions based on how much the denoiser benefits from training on them, rather than how well the current denoiser can reconstruct them.


To address this challenge, we introduce {\bf Variational Stackelberg Discrete Diffusion (VSDD)}, which formulates training as a \emph{variational Stackelberg game}. 
Instead of jointly optimizing the forward process and the denoiser with a shared objective, the \emph{leader} selects the corruption process while anticipating the follower's response, and the \emph{follower} optimizes the variational objective of the denoiser with the corruption process fixed. 
This separation discourages the forward process from reducing the training loss by adapting to the current denoiser. Instead, the leader is rewarded for selecting corruptions that improve the denoiser after the follower has learned from them.


The posterior of the learned forward process affects how the model resamples, or re-drafts, tokens at inference time. Effective re-drafting therefore requires capturing relationships between tokens. 
Uniform noise treats all substitutions equally, regardless of the structure learned by the denoiser. 
In contrast, re-drafting an almost complete sequence may require distinguishing between plausible, closely related alternatives. 
We therefore parameterize the forward process using token embeddings learned by the denoiser. These embeddings capture relationships between tokens, while the leader learns how those relationships should shape the noise process. Related tokens can thus receive higher or lower transition probabilities depending on which substitutions provide useful training signals for the denoiser.
We refer to this as a semantically aware noise process.


To optimize the forward noise process, the leader must anticipate the follower's response. 
Computing the best response would require fully optimizing the denoiser for each candidate noise process, which is prohibitively expensive. 
We therefore approximate it with a \emph{better response} using a single virtual gradient update on samples from the proposed noise process.
We measure how much this update improves the denoiser under a fixed reference noise process and use this improvement as the leader’s reward. 
The leader then optimizes the noise process using a score function estimator.

We evaluate our approach across three structurally different domains: molecular, text, and playlist generation. 
Across these domains, VSDD improves the primary generation quality metrics over a number of diffusion baselines with and without learnable noise. 

Our main contributions are:
\begin{enumerate}[leftmargin=0.99cm] 
    \item We formulate learning the forward noise process and the denoiser as a variational Stackelberg game, where corruptions are optimized based on how much the denoiser improves after learning from them rather than how well the current denoiser can reconstruct them.
    
    \item We introduce a semantically aware noise process parameterized by token embeddings learned by the denoiser, allowing the forward process to adapt to the learned structure of the domain. 
    
    \item We design an efficient training algorithm that approximates the follower’s best response with a single virtual gradient step and optimizes the noise process via a score function estimator.
    
    \item We show improved generation and re-drafting across molecular, text and playlist domains.
\end{enumerate}

\section{Related Work}
\label{sec:related}


\paragraph{Discrete diffusion models.}
\citet{Sohl_Dickstein} introduced diffusion based generative models for continuous data, later developed through DDPM~\citep{Ho_2020} and score based generative modeling~\citep{Song2020ScoreBasedGM}. 
\citet{austin2021structured} extended diffusion to discrete state spaces with D3PM, where the forward process is defined through categorical transition matrices, including uniform and absorbing corruption. 
\citet{Hoogeboom2021} proposed multinomial diffusion with argmax flows. 
More recent work focused on masked diffusion~\citep{NEURIPS2024_eb0b13cc, Shi2024} and score entropy based methods~\citep{Lou2024}. 
In particular,~\cite{NEURIPS2024_eb0b13cc} showed that masked diffusion with absorbing noise can substantially outperform uniform corruption for text generation. These results demonstrate that the choice of corruption process is consequential, but existing approaches typically prescribe this process in advance. Our work instead learns the structure of the corruption process according to how useful different corruptions are for training the denoiser.

\paragraph{Learned noise schedules and forward processes.}
In continuous diffusion, several works learn the noise schedule. \citet{nichol2021improved} learn a variance schedule through the variational bound, while  \citet{Kingma_2021} parameterize the signal to noise ratio via a neural network. \citet{dieleman2022continuousdiffusioncategoricaldata} show that the forward process in continuous diffusion is equivalent up to reparameterization, implying that learning the schedule affects the weighting of the training objective rather than the underlying generative model. 
In discrete diffusion, however, different noise processes change what the denoiser is trained to reconstruct. 
Learning this forward process in discrete diffusion is thus a modeling decision rather than only a training convenience.



For discrete diffusion, \citet{bartosh2026forwardlearned} recently proposed learning the forward process by jointly optimizing the  noise model and the denoiser via a score function based surrogate objective. 
This approach presents two challenges: the score function gradient can have high variance, and joint optimization may allow the forward process to reduce the training objective by adapting to the current denoiser rather than by providing useful training corruptions. 
Our work differs in how the forward process is optimized: VSDD evaluates corruptions based on how much the denoiser improves after learning from them, rather than how well they suit the current denoiser. 
To reduce the variance of the score function estimator, we evaluate multiple corruption samples and use a leave one out baseline.

\paragraph{Game-theoretic concepts in machine learning.}
Stackelberg games model sequential interactions where a leader chooses a strategy while anticipating a follower's response \citep{ConitzerSandholm}.
Stackelberg formulations have been studied in machine learning, including analyses of GANs \citep{Goodfellow_2014,Fiez_2020}. Our approach is related to MAML style meta learning \citep{maml_2017}, where outer optimization evaluates its objective after inner loop adaptation. 
Similarly, in VSDD the leader evaluates a corruption process via  the denoiser’s response after adapting to corruptions sampled from that process. 
Unlike these settings, the leader in VSDD parameterizes the forward process itself, learning which corruptions provide useful training signals for the denoiser.


\section{Notation and Preliminaries}\label{sec:model}
We let $V$ be the \textit{vocabulary size}, and let $\abar_t\in[0,1]$ be the cumulative \textit{retention schedule}.
Following standard notation in diffusion modeling \citep{austin2021structured}, we let $\alpha_t=\abar_t/\abar_{t-1}$ be the \textit{per-step retention rates} and $\beta_t=1-\alpha_t$ be the respective \textit{noise rates}.
Let $x_t=(x_t^1,\dots,x_t^L)$ be the sequence of $L$ tokens at time $t$, where each $x_t^\ell\in\{1,\dots,V\}$ is represented as a one-hot vector. 
One token is a \texttt{PAD} token that is never corrupted. 
We let $x_0$ be the \textit{clean} (uncorrupted) sequence.

In standard discrete diffusion, the forward noise process is fixed in advance: one specifies the retention schedule $\abar_t$, constructs the per-step token transition matrices $Q_t\in\mathbf{R}^{V\times V}$, and only trains the \textit{denoiser} $p_\theta(x_0 | x_t, t)$ against this fixed noise process $Q_t$ \citep{austin2021structured}. 
To this end, we let $\bar{Q}_t=\prod_{\tau=1}^t Q_t$ be the \textit{cumulative transition matrix} after $t$ steps.
The forward marginal probability distribution used to sample a noisy sequence $x_t$ is:
\begin{equation}\label{eq:categorical}
q(x_t\mid x_0)=\Cat\!\big(x_t | \,\Qbar_t^{\top}  \,x_0\big).
\end{equation}


In this setting, the denoiser $p_\theta(x_0 | x_t,t)$ is parameterized by a transformer with learnable parameters $\theta$. 
At inference, it takes a corrupted sequence $x_t$, the timestep $t$ and predicts the clean sequence $\hat{x}_0$.


We let \(E_\theta\in \mathbb{R}^{V \times d}\) denote the \textit{token embedding matrix} learned as part of the denoiser \(p_\theta(x_0 | x_t, t)\), with one \(d\)-dimensional embedding for each of the \(V\) vocabulary tokens.\footnote{Typically, the embedding layer is the first layer of the transformer used as denoiser.}
Finally, we use the standard ELBO loss $\mathcal{L}$ that includes the  per-step KL terms $\mathcal{L}_{\mathrm{KL}}(t)$ (for $t>1$) and the auxiliary reconstruction cross-entropy $\mathcal{L}_{\mathrm{CE}}$, over non-\texttt{PAD} positions:
\begin{equation}
\mathcal{L}=\sum_{t=2}^T\underbrace{\KL\!\big(q(x_{t-1}\mid x_t,x_0)\,\|\,p_\theta(x_{t-1}\mid x_t)\big)}_{\mathcal{L}_{\mathrm{KL}}(t)}\;\underbrace{-\log p_\theta(x_0\mid x_1)}_{\mathcal{L}_{\mathrm{CE}}}.
\label{eq:loss}
\end{equation}
Here, $q(x_{t-1}| x_t,x_0)$ is the true reverse posterior induced by the forward  process, and $p_\theta(x_{t-1}| x_t)$ is the corresponding learned posterior. 
In Section~\ref{sec:stackelberg}, we show how to compute these posteriors.

\section{Semantic-Aware Noise Process}
\label{sec:semantics}

As motivated in Section~\ref{sec:intro}, effective re-drafting may need to distinguish between plausible, closely related tokens  rather than treating all substitutions equally. 
We therefore construct a semantically aware noise whose transitions depend on the token representations learned by the denoiser. 
It provides the structure between tokens that we leverage to learn the forward process in Section \ref{sec:stackelberg}.

To parameterize the forward noise process by the token representations we let
\begin{align}\label{eq:leader_action}
    Q_{t,\phi} = \alpha_t\,I + (1-\alpha_t) M_{t,\phi}, \;\;\; t=1,...,T,
\end{align}
where $I$ is the identity matrix of rank $V$, and $M_{t,\phi}$ is the learnable transition matrix parameterized by $\phi$. 
Thus, at each step $t$, a token is retained with probability \(\alpha_t\), while with probability \(1-\alpha_t\) it is replaced according to \(M_{t,\phi}\).

To construct $M_{t,\phi}$, we define $s_{t, \phi}(i,j ) = e_i\, A_{t ,\phi} \,e_j^T$ as a learned similarity score between tokens $i$ and $j$. 
Here, $e_i$, $e_j\in\mathbf{R}^d$ are $L2$-normalized $i$th and $j$th rows from $E_\theta$ (Section \ref{sec:model}), and $A_{t ,\phi}\in \mathbb{R}^{d\times d}$ learns how relationships in the denoiser's embedding space should influence corruption at time $t$. 
In other words, the learned embeddings $e_i, e_j$ provide the underlying token structure, while \(A_{t,\phi}\) learns which aspects of this structure should determine likely substitutions.
For all non-pad tokens we define
\begin{align}\label{eq:M}
    M_{t,\phi}[i,j] = \mathbf{1}_{\{i\neq j\}} \frac{ \exp{s_{t, \phi}(i,j)}}{\sum_{\ell\neq i} \exp{s_{t,\phi}(i, \ell)}},
\end{align}
which ensures that $M_{t,\phi}$ and  $Q_{t,\phi}$ are row stochastic, meaning that every row defines a valid probability distribution over the next token. 
%
We let $\bar{Q}_{t,\phi} = \prod_{\tau=1}^t Q_{t,\phi}$ denote the corresponding cumulative transition matrix. 
Consequently, we  let the forward marginal be $q_{\phi}(x_t | x_0) = \Cat\big(x_t; \bar{Q}_{t,\phi}^T x_0\big)$ and its corresponding reverse posterior be $q_\phi (x_{t-1} | x_t, x_0)$.

A direct approach would be to add $\phi$ to the diffusion loss (Equation (\ref{eq:loss})) and to jointly optimize the forward process and the denoiser. 
However, this can lead to learning a degenerate noise process and a poor denoiser. 
To see this, observe that the KL term in Equation (\ref{eq:loss}) can decrease in two ways: (1) by improving the denoiser $p_\theta$ to better match the true posterior, or (2) by simplifying $q_\phi$ so that the posterior becomes trivial to match. 
Joint optimization permits the latter path: the forward process can adapt its off-diagonal transitions to the current denoiser, increasing the probabilities of substitutions that are already easy for \(p_\theta\), simplifying the posterior target. 
The denoising loss may therefore decrease without an improvement in generation quality. 
In Section~\ref{sec:stackelberg}, we address this by optimizing the noise process based on how the denoiser improves after learning from its corruptions.


\section{Variational Stackelberg Game}\label{sec:stackelberg}
%

To prevent the degenerate solutions discussed above, we formulate the optimization problem from a game theoretic perspective. 
In particular, we consider two players, a \emph{leader} and a \emph{follower}, interacting in a Stackelberg game fashion \citep{ConitzerSandholm}. 
The leader chooses an action \(\phi\) that parameterizes the forward corruption process \(Q_{t,\phi}\), while the follower observes this process and chooses an action \(\theta\) corresponding to the parameters of the denoiser \(p_\theta(x_0| x_t,t)\). 
Thus, the leader determines how training sequences are corrupted, while the follower learns to denoise sequences generated by that process. 
We now elaborate on the objectives of the follower and the leader and show how their interaction is used to jointly learn \(\phi\) and \(\theta\).

\paragraph{Follower.} 
The follower optimizes the standard diffusion loss from Equation~(\ref{eq:loss}) under the noise process selected by the leader. 
Importantly, as is common in Stackelberg games we assume that the follower cannot directly affect the action of the leader \citep{ConitzerSandholm}.
Therefore, when choosing its action $\theta$ the follower samples  noisy sequences $x_t$ from $q_{\sg, \phi}(x_t | x_0) = \Cat(x_t | \sg\{\bar{Q}_{t,\phi}\}^T x_0)$, where $\sg\{.\}$ is the stop-gradient operator.
This allows the follower to train on the corruption process selected by the leader while preventing gradients from the follower's objective from propagating to the leader parameters \(\phi\). 
Consequently, the follower's loss can be obtained from Equation~(\ref{eq:loss}) as follows:
\begin{align}\label{eq:follower_loos}
    \mathcal{L}^F(\theta |\phi) = (T-1) D_{KL}\big( q_{\sg, \phi}(x_{t-1}| x_t, x_0) || p_\theta(x_{t-1} | x_t) \big) - \log p_\theta (x_0| x_1),
\end{align}
Here, $q_{\sg, \phi}(x_{t-1}| x_t, x_0)$ is the reverse posterior corresponding to $q_{\sg, \phi}(x_t | x_0)$, and $p_\theta(x_{t-1} | x_t)$ is the denoiser's posterior, computed as
\begin{align}
    p_\theta(x_{t-1} | x_t) = \sum_{\hat{x}_0} p_\theta(\hat{x}_0 | x_t, t) q_{\sg, \phi}(x_{t-1}| x_t, \hat{x}_0),
\end{align}
Following \cite{austin2021structured}, this combines the denoiser's prediction over the clean sequence \(\hat x_0\) with the reverse posterior of the fixed corruption process to obtain the distribution over the previous diffusion state \(x_{t-1}\). 
Thus, the follower optimizes the standard discrete diffusion loss with the corruption process \(Q_{t,\phi}\) held fixed.


\paragraph{Leader.} 

The leader's goal is to find a corruption process $Q_{t,\phi}$ such that, after the follower optimizes the denoiser against this process (via Equation~(\ref{eq:follower_loos})), the quality of the generated sequences would be maximal. 
To this end, the leader needs to: (1) assess whether a candidate strategy $\phi$ improves the log-likelihood of cleaned data $\log p(\hat{x}_0)$, and (2) determine how to update $\phi$ while accounting for the follower's subsequent response $\theta$.

To address the first problem, we would ideally evaluate the resulting denoiser through the log likelihood of clean data, but optimizing this quantity directly is intractable. 
We therefore use an ELBO objective as a tractable surrogate. 
To ensure that different leader strategies are evaluated against the same corruption distribution, we introduce a fixed reference noise process \(Q_{\mathrm{ref}}\), and we let \(q_{\mathrm{ref}}(x_t | x_0)\) be its corresponding forward marginal distribution.\footnote{For example, we can use a uniform Categorical distribution as in standard uniform D3PM, Equation (\ref{eq:categorical}).} Using a fixed reference process provides a common evaluation distribution for comparing different leader strategies.

While the reference process determines the noise distribution used for evaluation, the current leader strategy \(\phi\) determines the reverse posterior $q_{\phi}(x_{t-1} | x_t, x_0)$ used to construct the reverse transition probabilities. 
As in the follower's case, we assume that the leader cannot affect the follower's action $\theta$ directly, and therefore, we freeze the gradient flow via the follower's embeddings, i.e., $\sg\{E_\theta\}$ when computing $\sg_\theta\{q_\phi\}$. 
The leader's reference ELBO objective results in
\begin{align}\label{eq:leader_elbo}
    \mathcal{L}_{\text{ELBO}}^L(\phi |\theta) = (T-1) D_{KL}\big(q_\text{ref}(x_{t-1}| x_t, x_0) || p_{\theta,\phi}(x_{t-1}|x_t) \big) - \log p_\theta(x_0 | x_1),
\end{align}
where
\begin{align}
    p_{\theta, \phi}(x_{t-1} | x_t) = \sum_{\hat{x}_0} p_\theta(\hat{x}_0 | x_t) \sg_\theta\{q_{\phi}(x_{t-1}| x_t, \hat{x}_0)\}.
\end{align}
The fixed reference process is important: the leader is evaluated against the same reference distribution as \(\phi\) changes, rather than against an evaluation distribution that changes with its own strategy.


In a Stackelberg game, the leader anticipates the follower's best response to its strategy \citep{ConitzerSandholm}. 
In our setting, computing the exact best response would require fully optimizing the denoiser for each candidate corruption process, which is prohibitively expensive. 
We therefore use a single gradient update as an approximate \textit{better response}. 
To this end, we sample a validation minibatch \(x_0\) and inject \(K\) different noise realizations into it to create \(K\) new minibatches \(x_t^{(k)}\), \(k=1,\ldots,K\). 
We let $d\theta_k = -\gamma\nabla_\theta \mathcal{L}^F(\theta;x_t^{(k)})$ denote the resulting virtual follower update for the \(k\)th validation minibatch, where \(\gamma\) is a step size hyperparameter. 
Thus, \(\theta+d\theta_k\) approximates how the follower would change after learning from that particular corruption realization.

We then compute the \textit{reward} for that noise realization as the relative improvement in the leader's reference loss:
\begin{align}
    R_k = \frac{\mathcal{L}_\text{ELBO}^L(\phi | \theta) - \mathcal{L}_\text{ELBO}^L(\phi | \theta+d\theta_k)}{\gamma |\mathcal{L}_\text{ELBO}^L(\phi | \theta)|}.
\end{align}
A positive reward therefore indicates that training on the sampled noise improves the follower under the fixed reference objective. 
This evaluates a corruption by the improvement it induces in the denoiser, rather than by how easily the current denoiser can reconstruct it.

These rewards allow the leader to increase the probability of noise realizations that improve the follower and decrease the probability of those that do not, using a score function estimator \citep{bartosh2026forwardlearned}. Because such estimators may have high variance, we normalize and clip the rewards:
\begin{align}\label{eq:reward_tilde}
    \Tilde{R}_k = \min \Big\{\max\big\{ \frac{R_k-B_k}{\sqrt{v + \epsilon}} , -c\big\} , c\Big\},\;\;\; \text{ where } B_k=\frac{1}{K-1}\sum_{j\neq k} R_j,
\end{align}
where \(B_k\) is a leave one out baseline, \(v\) is the running variance of the rewards, and \(c\) is the clipping threshold.
The final leader's objective is
\begin{align}\label{eq:leader_loss}
    \mathcal{L}^{L}(\phi)= -\frac{1}{K}\sum_{k=1}^{K}
\operatorname{sg}\{\widetilde{R}_k\}
\log q_\phi\!\left(x_t^{(k)} \mid x_0,t\right).
\end{align}
Minimizing this objective increases the probability of noise realizations with positive relative reward and decreases the probability of those with negative relative reward.



\paragraph{Stackelberg Game Dynamics.} 
As mentioned above, computing the exact best responses is prohibitively expensive.
We therefore organize training into blocks of $N$ steps.
At the start of each block, the leader uses a one-step virtual update to estimate the follower's expected better response and commits to a forward noise process $Q_{t,\phi}$.
The follower then performs \(N\) gradient updates while the noise process remains unchanged.

\begin{algorithm}[t]
\small
\caption{Variational Stackelberg Training for Discrete Diffusion}
\label{alg:stackelberg}
\begin{algorithmic}[1]
\REQUIRE Reference process $q_{\text{ref}}$, block size $N$, number of probes $K$; learning rates $\eta_\phi, \eta_\theta$; $\gamma$, $\lambda_T$
\STATE Initialize denoiser $p_\theta(x_0 | x_t, t)$ and noise kernel $Q_\phi$
\FOR{epoch $= 1, \dots M$}
    \STATE $\text{block\_steps} \leftarrow 0$
    \FOR{each minibatch $x_0$}
        \IF{$\text{block\_steps} = 0$}
            \STATE Build $Q_\phi, \Qbar_\phi$ 
        \ENDIF
        \STATE \textbf{Follower step:} $\theta \leftarrow \theta - \eta_\theta \nabla_\theta \mathcal{L}^{F}(\theta; \sg\{Q_{\phi}\})$ \hfill \COMMENT{Eq. ~(\ref{eq:follower_loos})}
        \STATE $\text{block\_steps} \leftarrow \text{block\_steps} + 1$
        \IF{$\text{block\_steps} = N$}
            \STATE Sample a validation batch $x_0^{\text{val}}$; compute $\mathcal{L}_\text{ELBO}^L(\phi | \theta)$ via Eq. (\ref{eq:leader_elbo}) 
            \FOR{$k = 1, \dots, K$} 
                \STATE $x_t^{(k)}\leftarrow$ Corrupt $x_0^\text{val}$ under $Q_\phi$
                \STATE Follower's better response: $\theta_{br} \leftarrow \theta - \gamma \nabla_\theta \mathcal{L}^{F}(\theta;x_t^{(k)})$
                \STATE Reward: $R_k \leftarrow \frac{\mathcal{L}_\text{ELBO}^L(\phi | \theta) - \mathcal{L}_\text{ELBO}^L(\phi | \theta_{br})}{\gamma |\mathcal{L}_\text{ELBO}^L(\phi | \theta)|}$
            \ENDFOR 
            \STATE Compute normalized rewards $\Tilde{R}_k$ via Eq. \ref{eq:reward_tilde} 
            \STATE $\mathcal{L}^{L}(\phi) \leftarrow \mathcal{L}^L(\phi)\;+\; \lambda_{T}\, D_{KL}\!\bigl[\bar{Q}_{T,\phi} \,\|\, \mathcal{U}\bigr]$ \hfill \COMMENT{Eq. \ref{eq:leader_loss} + terminal regularization} 
            \STATE \textbf{Leader step:} $\phi \leftarrow \phi - \eta_\phi \nabla_\phi \mathcal{L}^{L}(\phi)$ \hfill \COMMENT{Eq. (\ref{eq:leader_loss})}
            \STATE $\text{block\_steps} \leftarrow 0$
        \ENDIF
    \ENDFOR
\ENDFOR
\end{algorithmic}
\end{algorithm}

The full algorithm is summarized in Algorithm~\ref{alg:stackelberg}.
Here, we let $\eta_\theta$ and $\eta_\phi$ denote the learning rates of the follower and the leader, respectively. 
At the beginning of each training block, we fix the leader's forward corruption process $Q_\phi$ and the cumulative transition matrix  $\bar{Q}_\phi$,  which remain unchanged during the \(N\) follower updates (lines 5-6).
For each minibatch within the block, the follower performs a gradient update of $\theta$ by optimizing $\mathcal{L}^{F}$ via Equation~(\ref{eq:follower_loos}) (line 7).
At the end of the block, the leader samples a validation minibatch $x_0^{\text{val}}$ and computes its reference loss using Equation~(\ref{eq:leader_elbo}) (line 10).
The leader then creates \(K\) corrupted versions \(x_t^{(k)}\) of the same validation minibatch, computes a one-step better response for each corruption realization, and evaluates the corresponding reward (lines 11-14).
The rewards are then normalized using Equation \ref{eq:reward_tilde} (line 15).
In line 16, we compute the leader's loss via Equation (\ref{eq:leader_loss}) and add a terminal KL-regularization that encourages $\bar{Q}_{T,\phi}$ to match the prior (here, $\lambda_T$ is a regularization term).
Finally (line 17), the leader updates its action $\phi$ by minimizing its loss as in Equation (\ref{eq:leader_loss}).
Importantly,  this update is based on how much the sampled corruptions are predicted to improve the denoising process, rather than on how well they suit the current denoiser.

\section{Evaluation}\label{sec:evals}
We evaluate our approach on three domains that differ substantially in vocabulary size, structural constraints, and the nature of sequential dependencies: molecular, text, and playlist generation.
We compare the following five forward process designs, all sharing the same denoiser architecture and comparable training settings. More details on the datasets and  the setups are in Appendix~\ref{app:experiment_details}.

\begin{enumerate}[leftmargin=0.5cm] 
    \item \textbf{MDLM}~\citep{NEURIPS2024_eb0b13cc}. The standard absorbing forward process where tokens are independently replaced with a designated \texttt{[MASK]} token. 

    \item \textbf{D3PM-Uniform}~\citep{austin2021structured}. A uniform noise process in which corrupted tokens are replaced by tokens sampled uniformly at random. 
    The training objective combines a KL-term with a cross-entropy reconstruction loss following the standard discrete diffusion loss. 
    
    \item \textbf{Forward-Learned Discrete Diffusion (FLDD)} \citep{bartosh2026forwardlearned}. A non-Markov forward noise process that uses a REINFORCE surrogate objective to learn the forward process.

    \item \textbf{D3PM-Reinforce}. A learnable Markov forward process with the same parameterization as VSDD, in which \(\phi\) and \(\theta\) are jointly optimized using the score function surrogate objective of~\citet{bartosh2026forwardlearned}.

    \item \textbf{Variational Stackelberg Discrete Diffusion (VSDD)}. Our model, which uses the same learnable Markov forward process as D3PM-Reinforce but optimizes it through the Stackelberg dynamics (Algorithm~\ref{alg:stackelberg}). 
    The leader is updated every $N=100$ follower minibatch steps. 
\end{enumerate}

\subsection{Results}\label{sec:results}

\paragraph{Molecular generation.}
We use molecular data  
from a publicly available large-scale ZINC database \citep{Irwin2020ZINC20A}, represented as SMILES strings \citep{Weininger_Smiles}.
This domain imposes strict structural constraints: small syntactic errors can render an entire generated sequence invalid, making it particularly suitable for evaluating the re-drafting capabilities of our model. 
The small vocabulary of  64 tokens allows us to visually inspect the learned noise processes.
We evaluate the chemical validity of the generated molecular sequences, i.e., the fraction of generated SMILES strings that parse into valid molecules under RDKit~\citep{greg_landrum_2026_22140358}, as well as their uniqueness, novelty, and diversity.
All evaluations are computed over 15K generated samples per run.

\begin{figure}[t]
    \centering
    \begin{subfigure}[t]{0.25\textwidth}
        \centering
        \includegraphics[width=\linewidth]{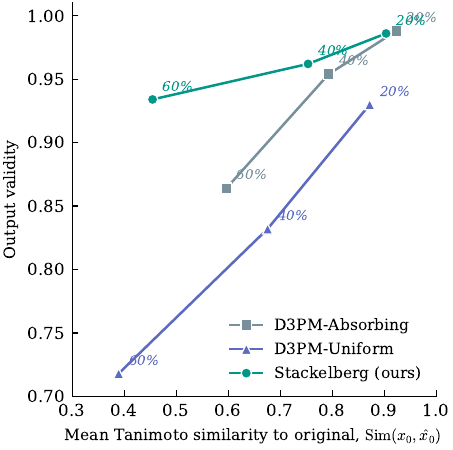}
        \caption{Validity/similarity after partial re-drafting.}
        \label{fig:redrafting}
    \end{subfigure}%
    \hspace{0.05\textwidth}%
    \begin{subfigure}[t]{0.60\textwidth}
        \centering
        \includegraphics[width=\linewidth]{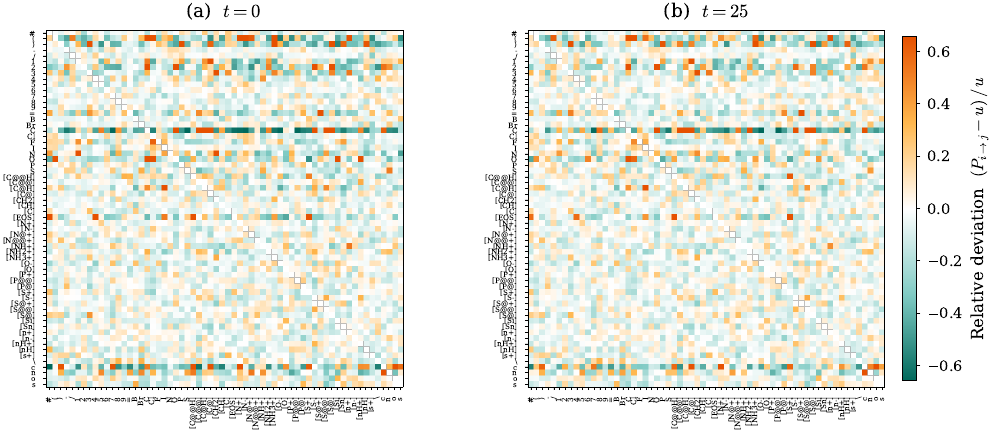}
        \caption{Learned transition matrices for the molecular domain. The color map is relative to the uniform noise.}
        \label{fig:learned_noise}
    \end{subfigure}

    \caption{Re-drafting and learned corruption in molecular generation.}
    \label{fig:Q_maps_cosine_bilinear}
\end{figure}

Table~\ref{tab:mol-validity} shows that 
VSDD achieves a validity rate of 84.3\%, substantially outperforming both the absorbing MDLM baseline (58.7\%) and D3PM-Uniform (48.1\%). 
We further inspect the learned off-diagonal transitions and observe several chemically interpretable patterns (Figure~\ref{fig:learned_noise}). 
At $t=25$, substitutions among halogens, such as \texttt{F}, \texttt{Cl}, \texttt{I}, are more likely than under uniform corruption. 
These halogens commonly attach to the rest of a molecule through a single chemical bond, so exchanging them can preserve the local bonding pattern.
The learned corruption also favors substitutions between the aromatic
atom tokens \texttt{c} and \texttt{n}.
These patterns suggest that the learned semantically aware noise process captures meaningful chemical relationships between tokens. 

\begin{figure}[t]
    \centering
    \begin{subfigure}[t]{0.28\textwidth}
        \centering
        \includegraphics[width=\linewidth]{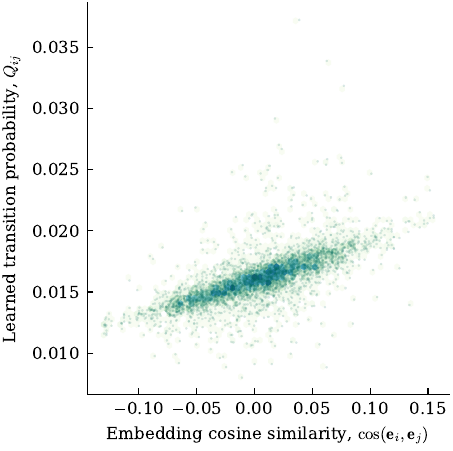}
        \caption{Molecular generation.}
        \label{fig:result-a}
    \end{subfigure}
    \hspace{0.07\textwidth}%
    \begin{subfigure}[t]{0.28\textwidth}
        \centering
        \includegraphics[width=\linewidth]{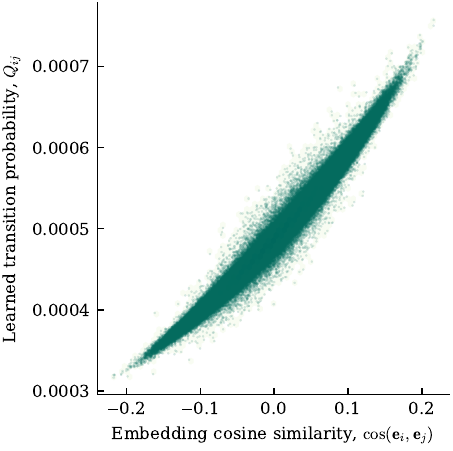}
        \caption{Text generation.}
        \label{fig:result-b}
    \end{subfigure}
    \hspace{0.07\textwidth}%
    \begin{subfigure}[t]{0.28\textwidth}
        \centering
        \includegraphics[width=\linewidth]{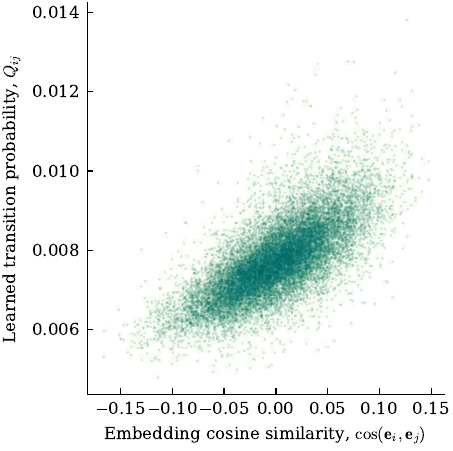}
        \caption{Playlist generation.}
        \label{fig:result-c}
    \end{subfigure}

    \caption{Correlation between the similarity of token embeddings and the transition probabilities.} 
    \label{fig:sim_prob_correlation}
\end{figure}

We further examine the relationship between the learned noise process and the denoiser's token embeddings. 
The global Spearman rank correlation between token-embedding cosine similarity and transition probability is $\rho = 0.70$ (Figure~\ref{fig:result-a}). 
Thus, tokens that are representationally similar under the denoiser's learned embeddings \(E_\theta\) tend to be more likely to be substituted for one another.
This correlation is absent by construction in D3PM-uniform and MDLM baselines. 

Finally, to inspect the re-drafting abilities of VSDD, we partially corrupt clean molecules by running the forward process for 20\%, 40\% and 60\% of the diffusion steps, then denoise them to reconstruct the originals.
Since different noise processes reach different corruption levels after the same number of steps, we plot the reconstruction validity against the mean Tanimoto similarity \citep{Bajusz2015WhyIT} between the corrupted molecules and clean ones. 
This provides a model-agnostic measure of the trade-off between the injected noise and reconstruction quality.
Figure~\ref{fig:redrafting} shows that VSDD achieves higher reconstruction validity than the baselines across all perturbation levels.

\begin{table}[t]
\centering
\caption{Chemical validity, uniqueness, novelty and diversity of generated molecules.}
\label{tab:mol-validity}
\begin{tabular}{lcccc}
\toprule
\textbf{Method} & \textbf{Validity (\%)} $\uparrow$ & \textbf{Uniqueness (\%)} $\uparrow$ & \textbf{Novelty (\%)} $\uparrow$& \textbf{Diversity} $\uparrow$ \\
\midrule 
MDLM  & $58.7$ & $100$ & $100$ & $0.8717$ \\
D3PM-Uniform    & $48.1$ & $100$ & $100$ & $0.8674$ \\
D3PM-Reinforce    & $62.0$ & $100$ & $100$ & $0.8727$\\
FLDD    & 3.0 & $100$ & $100$ & $0.9761$\\
VSDD (ours) & \textbf{84.3} & $100$ & $100$ & $0.8679$ \\
\bottomrule 
\end{tabular}
\end{table} 





\paragraph{Text Generation.}
We evaluate VSDD on the TinyStories dataset \citep{eldan2023tinystoriessmalllanguagemodels}, which allows us to study the learned forward process on natural language sequences with less rigid syntax and longer range semantic dependencies.
We preprocess data using a BPE tokenizer with a vocabulary of 2048 tokens and sequence length of 256 tokens.
For computational efficiency, we train all models on a random subset of 100K samples from the dataset.
We evaluate the generated stories using GPT-2 perplexity (PPL) and Distinct-N language diversity metric \citep{li2016}. 

VSDD more than halves the GPT-2 perplexity of D3PM-Uniform, reducing it from 236.33 to 106.71 (Table \ref{tbl:cosine_bilinear_text}).
This approaches MDLM's perplexity of 103.26 while achieving slightly higher Distinct-1 and Distinct-2 scores.  
Thus, VSDD substantially narrows the perplexity gap between uniform and masked diffusion while retaining the ability to revise previously generated tokens. 
These results show that the flexibility to re-draft need not incur the large perplexity penalty observed with unstructured uniform corruption.

In contrast, the jointly learned D3PM-Reinforce and FLDD baselines have significantly higher perplexity.
As discussed in Section~\ref{sec:semantics}, jointly optimizing the forward process and the denoiser can favor corruptions that are more predictable rather than informative. 
Minimizing the denoising loss $\mathcal{L}_{KL}(t)$ can therefore yield  nearly deterministic transitions (for example, $\Pr(\text{``was''}\rightarrow \text{``and''}) = 0.98$, and $\Pr(\text{``.''}\rightarrow \text{``to''})=0.99$;  
see Appendix \ref{app:degenerate_noise}).
In contrast, VSDD accounts for the follower's adaptation when updating the leader and favors transitions that improve generation quality under the reference noise process.
%
We further analyze VSDD's re-drafting capability in Appendix~\ref{app:redrafting_text}. 

\begin{table}[t]
    \centering
    \caption{Text and playlist generation results.}
    \label{tbl:text_and_playlists}

    \begin{subtable}[t]{0.49\textwidth}
        \vspace{0pt}
        \centering
        \setlength{\tabcolsep}{4pt}
        \begin{tabular}{@{}lccc@{}}
            \toprule
            Model & PPL $\downarrow$ & Dist-1 $\uparrow$ & Dist-2 $\uparrow$ \\
            \midrule
            MDLM           & $103.26$ & $0.1535$ & $0.5303$ \\
            D3PM-Uniform   & $236.33$ & $0.1362$ & $0.5718$ \\
            D3PM-Reinforce & $351.83$ & $0.2447$ & $0.7022$      \\
            FLDD           & $871.51$ & $0.2441$ & $0.7671$    \\
            VSDD (ours)    & $106.71$ & $0.1570$ & $0.5426$ \\
            \bottomrule
        \end{tabular}
        \caption{Perplexity and diversity of
                 generated stories.} 
        \label{tbl:cosine_bilinear_text}
    \end{subtable}%
    \hspace{0.01\textwidth}%
    \begin{subtable}[t]{0.49\textwidth}
        \vspace{0pt}
        \centering
        \setlength{\tabcolsep}{4pt}
        \begin{tabular}{@{}lcc@{}}
            \toprule
            Model & NDCG@3 $\uparrow$ & HitRate@3 $\uparrow$ \\
            \midrule
            MDLM           & $0.0882$          & $0.2020$ \\
            D3PM-Uniform & $0.0575$          & $0.1540$ \\
            D3PM-Reinforce & $0.0526$          & $0.1340$ \\
            FLDD & $0.0418$          & $0.1100$ \\
            VSDD (ours)   & $\mathbf{0.1077}$ & $\mathbf{0.2460}$ \\
            \bottomrule
        \end{tabular}
        \caption{NDCG@3 and HitRate@3 of
                 generated playlists.}
        \label{tbl:cosine_bilinear_playlists}
    \end{subtable}
\end{table}
 
\paragraph{Playlist generation.} 
We now use a real world playlist recommendation dataset comprising approximately 44K listening histories. 
Each history is a variable length sequence of discrete semantic IDs of the tracks the user interacted with.
Each semantic ID is represented by a triplet of tokens (see Appendix \ref{app:music_data}).
This domain allows us to evaluate our approach on sequences whose coherence is governed by behavioral and semantic relationships rather than an explicit formal grammar.

To evaluate our model, we hold out the final three tracks (9 tokens) of each validation playlist and perform conditional reverse diffusion. 
The prefix positions remain fixed to their ground-truth values throughout all reverse steps, and we only denoise the held-out suffix positions. 
A generated triplet is considered relevant if it exactly matches any of the three ground-truth held-out triplets. 
We report NDCG@3 and HitRate@3 on 500 playlists in Table~\ref{tbl:cosine_bilinear_playlists}. 
With only three relevant tracks among approximately 228K unique triplets, exact match prediction is challenging: a uniformly random prediction would have a hit probability of approximately $1.3 \times 10^{-5}$. 
VSDD achieves NDCG@3 of 0.108 and HitRate@3 of 0.246, outperforming all four baselines on both metrics.

In Appendix~\ref{app:redrafting_playlist}, we further study the re-drafting properties of VSDD by analyzing the model’s ability to repair targeted structural corruptions in playlists.

\subsection{Discussion}\label{sec:discussion}

Our evaluation highlights that both the choice of the corruption process and its optimization matter. 
VSDD outperforms the jointly learned D3PM-Reinforce and FLDD baselines on the primary metrics across all three domains. 
The comparison with D3PM-Reinforce is particularly informative because it uses the same noise parameterization but optimizes the forward process differently. 
These results support our central premise that useful corruptions should be selected by the improvement they induce after denoiser adaptation, rather than on how easily the current denoiser reconstructs them.  

The comparison with fixed corruption processes varies across domains. For molecular generation, VSDD substantially improves validity over both uniform and masked diffusion, while for playlist generation it improves both NDCG@3 and HitRate@3. For text generation, VSDD substantially reduces the perplexity of uniform diffusion and is comparable to masked diffusion. 

The learned noise and re-drafting analyses provide complementary evidence beyond generation quality. 
In molecules, the learned noise maps reveal pronounced, token-dependent structure, including chemically interpretable substitutions.
Furthermore, it achieves higher reconstruction validity across the evaluated partial-corruption levels. 
We observed similar results for the text and playlist generation tasks.
Together, these results show that the benefits of the learned corruption process extend beyond generation to re-drafting perturbed sequences across all three domains.


\section{Conclusion}\label{sec:conclusion}
We introduced Variational Stackelberg Discrete Diffusion (VSDD), a framework for learning a semantically aware forward noise process through the denoiser's response. The leader evaluates sampled corruptions by the improvement they induce after a virtual follower update, while the
follower learns to reverse the selected noise process.
Experiments on molecular, text, and playlist generation
show improvements over fixed and learnable forward noise process baselines. 
Our reconstruction and conditional-infilling experiments further show that these benefits extend to re-drafting, with VSDD effectively repairing controlled perturbations across all three domains.
Together, these findings suggest that effective discrete diffusion benefits from learning not only how to reverse corruption, but also which corruptions help the denoiser learn to revise.

\bibliographystyle{ACM-Reference-Format}

\bibliography{sample-base}

@String{Computing = "Computing" }

@String{Computer = "{IEEE} Computer" }

@String{Chelsea = "Chelsea" }

@inproceedings{NEURIPS2024_eb0b13cc,
  author    = {Sahoo, Subham Sekhar and Arriola, Marianne and Schiff, Yair
               and Gokaslan, Aaron and Marroquin, Edgar and Chiu, Justin T
               and Rush, Alexander and Kuleshov, Volodymyr},
  title     = {Simple and Effective Masked Diffusion Language Models},
  booktitle = {Advances in Neural Information Processing Systems},
  volume    = {37},
  year      = {2024},
  doi       = {10.52202/079017-4135},
  url       = {https://proceedings.neurips.cc/paper_files/paper/2024/hash/eb0b13cc515724ab8015bc978fdde0ad-Abstract-Conference.html}
}

@inproceedings{bartosh2026forwardlearned,
title     = {{Forward-Learned Discrete Diffusion: Learning How to Noise to Denoise Faster}},
author    = {Bartosh, Grigory and Pandeva, Teodora and Karmalkar, Sushrut and Zazo, Javier},
booktitle = {International Conference on Learning Representations},
year      = {2026},
eprint    = {2605.18204},
archivePrefix = {arXiv},
primaryClass  = {stat.ML},
doi       = {10.48550/arXiv.2605.18204},
url       = {https://openreview.net/forum?id=45EtKUdgbJ}
}

@inproceedings{nichol2021improved,
  title     = {Improved Denoising Diffusion Probabilistic Models},
  author    = {Nichol, Alexander Quinn and Dhariwal, Prafulla},
  booktitle = {Proceedings of the 38th International Conference on Machine Learning},
  pages     = {8162--8171},
  year      = {2021},
  editor    = {Meila, Marina and Zhang, Tong},
  volume    = {139},
  series    = {Proceedings of Machine Learning Research},
  month     = {18--24 July},
  publisher = {PMLR},
  url       = {https://proceedings.mlr.press/v139/nichol21a.html}
}

@inproceedings{austin2021structured,
  author    = {Austin, Jacob and
               Johnson, Daniel D. and
               Ho, Jonathan and
               Tarlow, Daniel and
               van den Berg, Rianne},
  title     = {Structured Denoising Diffusion Models in Discrete State-Spaces},
  booktitle = {Advances in Neural Information Processing Systems},
  editor    = {Ranzato, Marc'Aurelio and
               Beygelzimer, Alina and
               Dauphin, Yann and
               Liang, Percy S. and
               Wortman Vaughan, Jennifer},
  volume    = {34},
  pages     = {17981--17993},
  publisher = {Curran Associates, Inc.},
  year      = {2021},
  url       = {https://proceedings.neurips.cc/paper_files/paper/2021/file/958c530554f78bcd8e97125b70e6973d-Paper.pdf}
}

@article{BengioVJS15,
  author       = {Samy Bengio and
                  Oriol Vinyals and
                  Navdeep Jaitly and
                  Noam Shazeer},
  title        = {Scheduled Sampling for Sequence Prediction with Recurrent Neural Networks},
  journal      = {CoRR},
  volume       = {abs/1506.03099},
  year         = {2015},
  url          = {http://arxiv.org/abs/1506.03099},
  eprinttype   = {arXiv},
  eprint       = {1506.03099},
  bibsource    = {dblp computer science bibliography, https://dblp.org}
}

@article{Ranzato2015SequenceLT,
  title={Sequence Level Training with Recurrent Neural Networks},
  author={Marc'Aurelio Ranzato and Sumit Chopra and Michael Auli and Wojciech Zaremba},
  journal={CoRR},
  year={2015},
  volume={abs/1511.06732},
  url={https://api.semanticscholar.org/CorpusID:7147309}
}

@inproceedings{Lou2024,
author = {Lou, Aaron and Meng, Chenlin and Ermon, Stefano},
title = {Discrete diffusion modeling by estimating the ratios of the data distribution},
year = {2024},
publisher = {JMLR.org},
booktitle = {Proceedings of the 41st International Conference on Machine Learning},
articleno = {1333},
numpages = {30},
location = {Vienna, Austria},
series = {ICML'24}
}

@inproceedings{Shi2024,
author = {Shi, Jiaxin and Han, Kehang and Wang, Zhe and Doucet, Arnaud and Titsias, Michalis K.},
title = {Simplified and generalized masked diffusion for discrete data},
year = {2024},
isbn = {9798331314385},
publisher = {Curran Associates Inc.},
address = {Red Hook, NY, USA},
booktitle = {Proceedings of the 38th International Conference on Neural Information Processing Systems},
articleno = {3277},
numpages = {37},
location = {Vancouver, BC, Canada},
series = {NIPS '24}
}

@inproceedings{ConitzerSandholm,
author = {Conitzer, Vincent and Sandholm, Tuomas},
title = {Computing the optimal strategy to commit to},
year = {2006},
isbn = {1595932364},
publisher = {Association for Computing Machinery},
address = {New York, NY, USA},
url = {https://doi.org/10.1145/1134707.1134717},
doi = {10.1145/1134707.1134717},
booktitle = {Proceedings of the 7th ACM Conference on Electronic Commerce},
pages = {82–90},
numpages = {9},
location = {Ann Arbor, Michigan, USA},
series = {EC '06}
}

@article{Irwin2020ZINC20A,
  title={ZINC20 – A Free Ultra Large-Scale Chemical Database for Ligand Discovery},
  author={John J. Irwin and Khanh G. Tang and Jennifer Young and Chinzorig Dandarchuluun and Benjamin R. Wong and Munkhzul Khurelbaatar and Yurii S. Moroz and John W. Mayfield and Roger A. Sayle},
  journal={Journal of chemical information and modeling},
  year={2020},
  volume={60},
  pages={6065 - 6073},
  url={https://api.semanticscholar.org/CorpusID:226059428}
}

@article{Weininger_Smiles,
    author = {Weininger, David},
    title = {SMILES, a chemical language and information system. 1. Introduction to methodology and encoding rules},
    journal = {Journal of Chemical Information and Computer Sciences},
    volume = {28},
    number = {1},
    pages = {31-36},
    year = {2002},
    month = {05},
    issn = {0095-2338},
    doi = {10.1021/ci00057a005},
    url = {https://doi.org/10.1021/ci00057a005},
    eprint = {https://pubs.acs.org/jcisd8/article-pdf/28/1/31/10787984/ci00057a005.pdf},
}

@misc{eldan2023tinystoriessmalllanguagemodels,
      title={TinyStories: How Small Can Language Models Be and Still Speak Coherent English?}, 
      author={Ronen Eldan and Yuanzhi Li},
      year={2023},
      eprint={2305.07759},
      archivePrefix={arXiv},
      primaryClass={cs.CL},
      url={https://arxiv.org/abs/2305.07759}, 
}

@InProceedings{Sohl_Dickstein,
  title = 	 {Deep Unsupervised Learning using Nonequilibrium Thermodynamics},
  author = 	 {Sohl-Dickstein, Jascha and Weiss, Eric and Maheswaranathan, Niru and Ganguli, Surya},
  booktitle = 	 {Proceedings of the 32nd International Conference on Machine Learning},
  pages = 	 {2256--2265},
  year = 	 {2015},
  editor = 	 {Bach, Francis and Blei, David},
  volume = 	 {37},
  series = 	 {Proceedings of Machine Learning Research},
  address = 	 {Lille, France},
  month = 	 {07--09 Jul},
  publisher =    {PMLR},
  url = 	 {https://proceedings.mlr.press/v37/sohl-dickstein15.html}
}

@inproceedings{Ho_2020,
author = {Ho, Jonathan and Jain, Ajay and Abbeel, Pieter},
title = {Denoising diffusion probabilistic models},
year = {2020},
isbn = {9781713829546},
publisher = {Curran Associates Inc.},
address = {Red Hook, NY, USA},
booktitle = {Proceedings of the 34th International Conference on Neural Information Processing Systems},
articleno = {574},
numpages = {12},
location = {Vancouver, BC, Canada},
series = {NIPS '20}
}

@inproceedings{Kingma_2021,
author = {Kingma, Diederik P. and Salimans, Tim and Poole, Ben and Ho, Jonathan},
title = {Variational diffusion models},
year = {2021},
isbn = {9781713845393},
publisher = {Curran Associates Inc.},
address = {Red Hook, NY, USA},
booktitle = {Proceedings of the 35th International Conference on Neural Information Processing Systems},
articleno = {1660},
numpages = {12},
series = {NIPS '21}
}

@misc{dieleman2022continuousdiffusioncategoricaldata,
      title={Continuous diffusion for categorical data}, 
      author={Sander Dieleman and Laurent Sartran and Arman Roshannai and Nikolay Savinov and Yaroslav Ganin and Pierre H. Richemond and Arnaud Doucet and Robin Strudel and Chris Dyer and Conor Durkan and Curtis Hawthorne and Rémi Leblond and Will Grathwohl and Jonas Adler},
      year={2022},
      eprint={2211.15089},
      archivePrefix={arXiv},
      primaryClass={cs.CL},
      url={https://arxiv.org/abs/2211.15089}, 
}

@inproceedings{Goodfellow_2014,
author = {Goodfellow, Ian J. and Pouget-Abadie, Jean and Mirza, Mehdi and Xu, Bing and Warde-Farley, David and Ozair, Sherjil and Courville, Aaron and Bengio, Yoshua},
title = {Generative adversarial nets},
year = {2014},
publisher = {MIT Press},
address = {Cambridge, MA, USA},
booktitle = {Proceedings of the 28th International Conference on Neural Information Processing Systems - Volume 2},
pages = {2672–2680},
numpages = {9},
location = {Montreal, Canada},
series = {NIPS'14}
}

@InProceedings{maml_2017,
  title = 	 {Model-Agnostic Meta-Learning for Fast Adaptation of Deep Networks},
  author =       {Chelsea Finn and Pieter Abbeel and Sergey Levine},
  booktitle = 	 {Proceedings of the 34th International Conference on Machine Learning},
  pages = 	 {1126--1135},
  year = 	 {2017},
  editor = 	 {Precup, Doina and Teh, Yee Whye},
  volume = 	 {70},
  series = 	 {Proceedings of Machine Learning Research},
  month = 	 {06--11 Aug},
  publisher =    {PMLR},
  url = 	 {https://proceedings.mlr.press/v70/finn17a.html},
}

@InProceedings{Fiez_2020,
  title = 	 {Implicit Learning Dynamics in Stackelberg Games: Equilibria Characterization, Convergence Analysis, and Empirical Study},
  author =       {Fiez, Tanner and Chasnov, Benjamin and Ratliff, Lillian},
  booktitle = 	 {Proceedings of the 37th International Conference on Machine Learning},
  pages = 	 {3133--3144},
  year = 	 {2020},
  editor = 	 {III, Hal Daumé and Singh, Aarti},
  volume = 	 {119},
  series = 	 {Proceedings of Machine Learning Research},
  month = 	 {13--18 Jul},
  publisher =    {PMLR},
  url = 	 {https://proceedings.mlr.press/v119/fiez20a.html},
}

@inproceedings{Moor_CIKM_2023,
author = {Moor, Dmitrii and Yuan, Yi and Mehrotra, Rishabh and Dai, Zhenwen and Lalmas, Mounia},
title = {Exploiting Sequential Music Preferences via Optimisation-Based Sequencing},
year = {2023},
isbn = {9798400701245},
publisher = {Association for Computing Machinery},
address = {New York, NY, USA},
url = {https://doi.org/10.1145/3583780.3615476},
doi = {10.1145/3583780.3615476},
booktitle = {Proceedings of the 32nd ACM International Conference on Information and Knowledge Management},
pages = {4759–4765},
numpages = {7},
location = {Birmingham, United Kingdom},
series = {CIKM '23}
}

@article{Bajusz2015WhyIT,
  title={Why is Tanimoto index an appropriate choice for fingerprint-based similarity calculations?},
  author={D{\'a}vid Bajusz and Anita R{\'a}cz and K{\'a}roly H{\'e}berger},
  journal={Journal of Cheminformatics},
  year={2015},
  volume={7},
  url={https://api.semanticscholar.org/CorpusID:1221969}
}

@article{Song2020ScoreBasedGM,
  title={Score-Based Generative Modeling through Stochastic Differential Equations},
  author={Yang Song and Jascha Narain Sohl-Dickstein and Diederik P. Kingma and Abhishek Kumar and Stefano Ermon and Ben Poole},
  journal={ArXiv},
  year={2020},
  volume={abs/2011.13456},
  url={https://api.semanticscholar.org/CorpusID:227209335}
}

@inproceedings{li2016,
    title = "A Diversity-Promoting Objective Function for Neural Conversation Models",
    author = "Li, Jiwei  and
      Galley, Michel  and
      Brockett, Chris  and
      Gao, Jianfeng  and
      Dolan, Bill",
    editor = "Knight, Kevin  and
      Nenkova, Ani  and
      Rambow, Owen",
    booktitle = "Proceedings of the 2016 Conference of the North {A}merican Chapter of the Association for Computational Linguistics: Human Language Technologies",
    month = jun,
    year = "2016",
    address = "San Diego, California",
    publisher = "Association for Computational Linguistics",
    url = "https://aclanthology.org/N16-1014/",
    doi = "10.18653/v1/N16-1014",
    pages = "110--119"
}

@software{greg_landrum_2026_22140358,
  author       = {Greg Landrum and
                  Paolo Tosco and
                  Ricardo Rodriguez and
                  Brian Kelley and
                  David Cosgrove and
                  Riccardo Vianello and
                  sriniker and
                  Peter Gedeck and
                  Gareth Jones and
                  Dan Nealschneider and
                  Eisuke Kawashima and
                  NadineSchneider and
                  tadhurst-cdd and
                  Andrew Dalke and
                  Niels Maeder and
                  Matt Swain and
                  Yakov Pechersky and
                  Brian Cole and
                  Kevin Boyd and
                  Samo Turk and
                  Aleksandr Savelev and
                  Rachel Walker and
                  Alain Vaucher and
                  Maciej Wójcikowski and
                  Hussein Faara and
                  Ichiru Take and
                  Vincent F. Scalfani and
                  Steven Kearnes and
                  Kazuya Ujihara and
                  Daniel Probst},
  title        = {rdkit/rdkit: 2026\_03\_6 (Q1 2026) Release},
  month        = aug,
  year         = 2026,
  publisher    = {Zenodo},
  version      = {Release\_2026\_03\_6},
  doi          = {10.5281/zenodo.22140358},
  url          = {https://doi.org/10.5281/zenodo.22140358},
}

@inproceedings{Hoogeboom2021,
author = {Hoogeboom, Emiel and Nielsen, Didrik and Jaini, Priyank and Forr{\'e}, Patrick and Welling, Max},
title = {Argmax flows and multinomial diffusion: learning categorical distributions},
year = {2021},
isbn = {9781713845393},
publisher = {Curran Associates Inc.},
address = {Red Hook, NY, USA},
booktitle = {Proceedings of the 35th International Conference on Neural Information Processing Systems},
articleno = {953},
numpages = {12},
series = {NIPS '21}
}

@inproceedings{NIPS2017_Vaswani,
 author = {Vaswani, Ashish and Shazeer, Noam and Parmar, Niki and Uszkoreit, Jakob and Jones, Llion and Gomez, Aidan N and Kaiser, \L ukasz and Polosukhin, Illia},
 booktitle = {Advances in Neural Information Processing Systems},
 editor = {I. Guyon and U. Von Luxburg and S. Bengio and H. Wallach and R. Fergus and S. Vishwanathan and R. Garnett},
 pages = {},
 publisher = {Curran Associates, Inc.},
 title = {Attention is All you Need},
 url = {https://proceedings.neurips.cc/paper/2017/file/3f5ee243547dee91fbd053c1c4a845aa-Paper.pdf},
 volume = {30},
 year = {2017}
}

@ArtifactSoftware{R,
    title = {R: A Language and Environment for Statistical Computing},
    author = {{R Core Team}},
    organization = {R Foundation for Statistical Computing},
    address = {Vienna, Austria},
    year = {2019},
    url = {https://www.R-project.org/},
}








  

\newpage
\appendix
\section{Appendix}\label{app:derivations}

\section{Additional Experimental Details}\label{app:experiment_details}

In all learnable Markov noise models in our experiments (i.e., D3PM-Reinforce and VSDD, see Section \ref{sec:evals}) we rely on the same architecture of the forward noise process $Q_{t,\phi}$ and the learnable noise kernel.
Specifically, the off-diagonal transition kernel $M_{t,\phi}$ is parameterised by a time-conditioned bilinear scoring module that
computes token-to-token affinities according to Equation (\ref{eq:M}), and $A(t) \in \mathbb{R}^{d \times d}$ is produced by a two-layer MLP with SiLU activations operating on sinusoidal time features.
Rows of the score matrix are softmax-normalised, giving a valid row-stochastic off-diagonal kernel $M_{t, \phi}$.
The full single-step transition matrix $Q_{t, \phi}$ is then computed according to Equation (\ref{eq:leader_action}), and the cumulative product $\bar{Q}_{t,\phi} = \prod_{\tau=1}^{t} Q_{\tau,\phi}$ is computed sequentially. 
A terminal-uniform regularizer encourages $\bar{Q}_T$ to converge to a uniform stationary distribution over valid tokens.
 
\subsection{Molecular Domain}\label{app:mol_data}
\paragraph{Data \& Vocabulary.}
In the molecular domain, we randomly sample 1M molecules and use a 90/10 train/validation split.
We tokenize SMILES sequences provided in the dataset at the atom level using a custom tokenizer that handles bracket atoms (e.g., \texttt{[C@@H]}, \texttt{[NH+]}), two-letter elements (e.g., \texttt{Cl}, \texttt{Br}), single-letter atoms, and structural characters (bonds, branches, ring closures). This yields a compact vocabulary of $|\mathcal{V}|=64$ tokens and a maximum sequence length of $L=134$. An \texttt{[EOS]} token terminates each molecule; sequences shorter than $L$ are right-padded with \texttt{[PAD]}.

\paragraph{Denoiser architecture.}
All baselines as well as our proposed model use the same denoising backbone: a bi-directional Transformer encoder~\citep{NIPS2017_Vaswani} with the embedding dimensionality $d=512$, ten layers, 32 attention heads, and sinusoidal time embeddings injected additively into the input representation. 
The diffusion horizon is $T=50$ steps with a linear noise schedule $\bar{\alpha}_t$ decreasing from $1$ to $0$. Models are trained for 50 epochs with the Adam optimizer (learning rate $10^{-4}$, batch size 512) on two A100 GPUs.  

\subsection{Text Generation Domain}\label{app:text}

\paragraph{Data \& Vocabulary.} We evaluate on the TinyStories dataset \citep{eldan2023tinystoriessmalllanguagemodels}, tokenised with a byte-pair encoding (BPE) tokeniser trained on the full training split, yielding a vocabulary of $V= 2048$ tokens and a maximum sequence length of $L = 256$. 
Three special tokens are
reserved: [PAD] for right-padding sequences shorter than $L$, [EOS] for end-of-sequence, and [MASK] (used only by the masked-diffusion baselines).  
We train on 100k sequences randomly selected from the training split and validate on a held-out set of 1,000 sequences.

\paragraph{Denoiser architecture.} All models share the same denoising network $p_\theta(x_0 | x_t, t)$: a bi-directional Transformer encoder with the embedding dimension $d = 512$, ten layers, sixteen attention heads, and sinusoidal time embeddings added to the input representation. 
The model takes noisy tokens $x_t$ and a time embedding as input and outputs logits over the vocabulary at every sequence position. 
We perform training on multiple A100 80 GB GPUs with a batch size of 256.
The diffusion horizon has $T=50$ steps with a linear noise schedule $\bar\alpha_t$ decreasing from 1 to 0. 
The denoiser is trained with Adam optimizer with the learning rate $\eta_\theta = 10^{-4}$, weight decay $10^{-4}$, and gradient clipping at 1.0. 
The kernel parameters $\phi$ are optimized with a separate Adam instance with the learning rate $\eta_\phi=2 \times 10^{-4}$.

\subsection{Playlist Domain}\label{app:music_data}

\paragraph{Data \& Vocabulary.} We evaluate our model on a proprietary playlist dataset comprising listening sequences derived from over one million unique music tracks, each represented by an 80-dimensional embedding vector. 
The dataset consists of 44,042 listening histories (36,765 training, 7,277 validation), where each history is a variable-length sequence of tracks. 
To obtain a discrete vocabulary suitable for our model, we map each track's 80-dimensional embedding vector to a \textit{semantic ID} (SID) via locality-sensitive hashing (LSH): 21 random hyperplanes are partitioned into 3 levels of 7 bits each, yielding $2^7 = 128$ buckets per level plus 3 special tokens (PAD, MASK, EOS) for a vocabulary of $V = 131$. 
Each track is thus represented by a triplet $(t_0, t_1, t_2)$ of \textit{semantic tokens}, and the resulting corpus contains 228,132 unique triplets. 
Sequences are truncated or padded to a fixed length of $L = 150$ semantic tokens. 

\paragraph{Denoiser architecture.}
The denoiser architecture for playlist generation is similar to the other two tasks.
We let the dimensionality of the token embeddings be $d = 512$, and we use ten layers with 16 attention heads, trained for 50 epochs with batch size 28 and the Adam optimizer (follower $\text{lr} = 10^{-4}$, leader $\text{lr} = 2 \times 10^{-4}$). 
The diffusion process uses $T = 50$ steps.



\section{Re-Drafting Analysis}
In this section, we expand our analysis of the \textit{re-drafting} (conditional infilling) capabilities of VSDD for the text and playlist generation domains.
Conceptually, such an analysis is similar to the qualitative analysis that we presented for the molecular generation domain (see Section \ref{sec:results}, last paragraph).
However, as both the playlist and the text generation tasks have significantly larger vocabularies, the visual analysis of the learned forward noise process (Figure \ref{fig:learned_noise}) is not straightforward in these domains.
Instead, we construct a number of rule-based re-drafting tests, and we analyse the performance of our model on those tests.

\subsection{Text Generation}\label{app:redrafting_text}
In the text generation domain, we evaluate three discrete diffusion models trained on the TinyStories dataset. 
Re-drafting tests whether a model, given a story with a localised corruption, can restore the original content by conditioning on the uncorrupted context. 
This tests the model's ability to maintain a coherent narrative and factual agreement across the sequence.

To this end, we compare our VSDD model that learns the forward noise process against two baselines that rely on a fixed (pre-computed) forward noise process, namely, MDLM and D3PM-Uniform.
We then apply \textit{repaint-style} conditional infilling, i.e., given a sequence with a designated repainting mask, we use the model to regenerate only the masked positions while keeping the context positions fixed.

In the VSDD and D3PM-Uniform experiments, we initialise the positions marked with the repainting mask with uniform random noise over the valid tokens. 
The full reverse diffusion is then run for $T$ steps.
At each step, the model predicts $p_\theta(x_0 | x_t, t)$ and computes the posterior sampling probabilities $p(x_{t-1} | x_t, x_0)$.
It then samples the new tokens from this posterior. 
Only the masked positions are updated; context positions are held fixed.
For the MDLM model the edit positions are set to the absorbing mask token [MASK]. 
Only currently-masked non-pad positions are updated at each step while the context positions remain unchanged.
All experiments use five independent trials per corruption per model.

\paragraph{Experiment 1: Entity Swap.} 
For the first experiment, we construct five templated stories with repeated entities (names and objects). 
Each story contains 3--5 occurrences of a target entity. 
We corrupt exactly one occurrence by replacing its BPE token (see Appendix \ref{app:text}) with a different entity token, creating a single-token inconsistency. 
All entity names (``Lily", ``Tom", ``Max", ``Sam", ``Ben") and objects (``ball", ``dog", ``car", ``cat", ``train") are verified to be single BPE tokens, ensuring the corruption is a clean one-token swap with no sequence-length change.

Observe that in this setting the absorbing-state mechanism of MDLM is particularly well-suited for single-token infilling.
Indeed, MDLM is directly optimized for this specific task of unmasking a single masked token at the corrupted position with full surrounding context.
Consequently, its training objective is precisely to predict the clean token at masked positions.
This provides an ideal solution for entity swap, and we are interested in analysing how close the rest of the models can approach MDLM on this task.

The five corruption instances are in Table \ref{tab:story-corruptions}, and we provide an example of such a corruption in Table \ref{tab:corruption_1}.
For each story we run five corruption trials, and average the resulting metrics across the trials and stories.
We measure the exact repair rate as a fraction of the corrupted tokens restored to the original value (averaged over all trials) and consistency as the fraction of all entity occurrences that agree with the most-common value after infilling. 
A score of 1.0 means all instances of the entity hold the same token.
Table \ref{tab:aggregate-results} contains the aggregated results.

From Table \ref{tab:aggregate-results} we see that VSDD closely approaches the ``ideal" performance of MDLM, leaving a substantial gap with D3PM-Uniform.
These results indicate that even though VSDD was not trained to perform well on the entity swap task, it manages to efficiently re-draft the corrupted parts of the sequence. 
Despite the small performance gap with MDLM, below we illustrate that the infilled stories are still structurally and semantically consistent.

\begin{table}[t]
    \centering
    \caption{Synthetic story corruptions. For each story, an entity is replaced at a selected occurrence.}
    \label{tab:story-corruptions}
    
    \setlength{\tabcolsep}{9pt}
    \renewcommand{\arraystretch}{1.15}

    \begin{tabular}{@{}l l c c@{}}
        \toprule
        \textbf{Story name} &
        \textbf{Corruption} &
        \textbf{\# occurrences} &
        \textbf{Corrupted occurrence} \\
        \midrule
        Lily and red ball & Lily $\rightarrow$ Tom   & 4 & 2 \\
        Tom and Max       & Max $\rightarrow$ Sam    & 5 & 1 \\
        Sam and blue car  & car $\rightarrow$ dog    & 4 & 3 \\
        Lily and cat      & cat $\rightarrow$ dog    & 5 & 2 \\
        Ben and toy train & train $\rightarrow$ ball & 5 & 3 \\
        \bottomrule
    \end{tabular}
\end{table}

\begin{table}[t]
    \centering
    \caption{Aggregate repair performance.
    Results report the mean and the standard deviation across corruption types.}
    \label{tab:aggregate-results}

    \setlength{\tabcolsep}{10pt}
    \renewcommand{\arraystretch}{1.12}

    \begin{tabular}{@{}lccc@{}}
        \toprule
        \textbf{Model} &
        \textbf{Exact Repair} $\uparrow$ &
        \textbf{Consistency} $\uparrow$ \\
        \midrule
        VSDD  & $0.880 \pm 0.160$ & $0.974 \pm 0.033$  \\
        D3PM-Uniform & $0.360 \pm 0.265$ & $0.860 \pm 0.065$  \\
        MDLM         & $0.960 \pm 0.080$ & $0.992 \pm 0.016$  \\
        \bottomrule
    \end{tabular}
\end{table}


\begin{table}[t]
    \centering
    \caption{The 2nd occurrence of ``Lily'' is corrupted to ``Tom'', each model attempts to recover the original token.}
    \label{tab:corruption_1}

    \small
    \setlength{\tabcolsep}{6pt}
    \renewcommand{\arraystretch}{1.15}

    \begin{tabular}{
        @{}
        p{0.16\linewidth}
        p{0.66\linewidth}
        p{0.10\linewidth}
        @{}
    }
        \toprule

        \multicolumn{3}{@{}l@{}}{
            \textbf{Example 1: Lily $\rightarrow$ Tom}
            \hspace{1.2em}
            \textit{(2nd occurrence)}
        } \\[3pt]

        \midrule

        \multicolumn{3}{@{}p{\linewidth}@{}}{
            \textbf{Original}\quad
            Once upon a time there was a little girl named Lily.
            \textbf{Lily} had a big red ball.
            She loved to play with the red ball every day.
            One day Lily took the red ball to the park.
            Lily was very happy.
        } \\[6pt]

        \multicolumn{3}{@{}p{\linewidth}@{}}{
            \textbf{Corrupted}\quad
            Once upon a time there was a little girl named Lily.
            \textbf{Tom} had a big red ball.
            She loved to play with the red ball every day.
            One day Lily took the red ball to the park.
            Lily was very happy.
        } \\[3pt]

        \midrule


        \textbf{Model}
        &
        \textbf{Infilled text}
        &
        \textbf{Repair?}
        \\

        \midrule

        VSDD
        &
        \ldots named Lily. \textbf{Lily} had a big red ball.
        She loved \ldots
        &
        Yes
        \\[4pt]

        D3PM-Uniform
        &
        \ldots named Lily. \textbf{She} had a big red ball.
        She loved \ldots

        \vspace{1pt}
        {\footnotesize\itshape
        Generates ``She'' instead of ``Lily''.}
        &
        No
        \\[4pt]

        MDLM
        &
        \ldots named Lily. \textbf{Lily} had a big red ball.
        She loved \ldots
        &
        Yes
        \\

        \bottomrule
    \end{tabular}
\end{table}

In our first example (Table \ref{tab:corruption_1}), D3PM-Uniform never recovers "Lily" across all 5 trials (repair = 0.0); it consistently generates the pronoun "She", which is grammatically plausible but breaks the entity pattern. 
VSDD recovers "Lily" in 4 out of 5 trials (repair = 0.8).

\begin{table}[t]
    \centering
    \caption{The 1st occurrence of ``Max'' is corrupted to ``Sam''.}
    \label{tab:corruption_2}

    \small
    \setlength{\tabcolsep}{6pt}
    \renewcommand{\arraystretch}{1.15}

    \begin{tabular}{
        @{}
        p{0.16\linewidth}
        p{0.66\linewidth}
        p{0.10\linewidth}
        @{}
    }
        \toprule

        \multicolumn{3}{@{}l@{}}{
            \textbf{Example 2: Max $\rightarrow$ Sam}
            \hspace{1.2em}
            \textit{(1st occurrence)}
        } \\[3pt]

        \midrule


        \multicolumn{3}{@{}p{\linewidth}@{}}{
            \textbf{Original}\quad
            Tom had a small dog named \textbf{Max}.
            Tom and Max liked to play in the park.
            Every morning Tom took Max for a long walk.
            Max was a good dog and Tom loved Max very much.
        } \\[6pt]

        \multicolumn{3}{@{}p{\linewidth}@{}}{
            \textbf{Corrupted}\quad
            Tom had a small dog named \textbf{Sam}.
            Tom and Max liked to play in the park.
            Every morning Tom took Max for a long walk.
            Max was a good dog and Tom loved Max very much.
        } \\[3pt]

        \midrule


        \textbf{Model}
        &
        \textbf{Infilled text}
        &
        \textbf{Repair?}
        \\

        \midrule

        VSDD
        &
        \ldots dog named \textbf{Max}.
        Tom and Max liked \ldots

        \vspace{1pt}
        {\footnotesize\itshape
        Successful in 3/5 trials.}
        &
        Yes
        \\[4pt]

        D3PM-Uniform
        &
        \ldots dog named \textbf{Max}.
        Tom and Max liked \ldots

        \vspace{1pt}
        {\footnotesize\itshape
        Successful in 1/5 trials.}
        &
        Rarely
        \\[4pt]

        MDLM
        &
        \ldots dog named \textbf{Tom}.
        Tom and Max liked \ldots

        \vspace{1pt}
        {\footnotesize\itshape
        Repairs in 4/5 trials, but occasionally generates
        ``Tom'' instead of ``Max''.}
        &
        Partial
        \\

        \bottomrule
    \end{tabular}
\end{table}

The second example (Table \ref{tab:corruption_2}) is the hardest corruption: "Max" appears 5 times but the corrupted position is the first occurrence, where the model must choose between two plausible names ("Max" from later context vs. "Sam" or "Tom" from the immediate sentence). 
MDLM occasionally generates "Tom" (the human character's name) instead of "Max", showing a consistency failure despite high overall repair rate.

\begin{table}[t]
    \centering
    \caption{
    The third occurrence of ``car'' is corrupted to ``dog''.}
    \label{tab:corruption_3}

    \small
    \setlength{\tabcolsep}{6pt}
    \renewcommand{\arraystretch}{1.15}

    \begin{tabular}{
        @{}
        p{0.16\linewidth}
        p{0.66\linewidth}
        p{0.10\linewidth}
        @{}
    }
        \toprule

        \multicolumn{3}{@{}l@{}}{
            \textbf{Example 3: car $\rightarrow$ dog}
            \hspace{1.2em}
            \textit{(3rd occurrence)}
        } \\[3pt]

        \midrule


        \multicolumn{3}{@{}p{\linewidth}@{}}{
            \textbf{Original}\quad
            There was a little boy named Sam.
            Sam had a blue car.
            Sam liked to play with his blue car.
            One day Sam lost his blue \textbf{car}.
            Sam was sad but then he found the blue car under the bed.
        } \\[6pt]

        \multicolumn{3}{@{}p{\linewidth}@{}}{
            \textbf{Corrupted}\quad
            There was a little boy named Sam.
            Sam had a blue car.
            Sam liked to play with his blue car.
            One day Sam lost his blue \textbf{dog}.
            Sam was sad but then he found the blue car under the bed.
        } \\[3pt]

        \midrule


        \textbf{Model}
        &
        \textbf{Infilled text}
        &
        \textbf{Repair?}
        \\

        \midrule

        VSDD
        &
        \ldots lost his blue \textbf{car}.
        Sam was sad \ldots

        \vspace{1pt}
        {\footnotesize\itshape
        Successful in 5/5 trials.}
        &
        Yes
        \\[4pt]

        D3PM-Uniform
        &
        \ldots lost his blue \textbf{car}.
        Sam was sad \ldots

        \vspace{1pt}
        {\footnotesize\itshape
        Successful in 4/5 trials.}
        &
        Mostly
        \\[4pt]

        MDLM
        &
        \ldots lost his blue \textbf{car}.
        Sam was sad \ldots

        \vspace{1pt}
        {\footnotesize\itshape
        Successful in 5/5 trials.}
        &
        Yes
        \\

        \bottomrule
    \end{tabular}
\end{table}

With strong contextual cues (three other occurrences of "blue car" surrounding the corruption), all models perform well (Table \ref{tab:corruption_3}).  
In this case, even D3PM-Uniform achieves 80\% repair, demonstrating that sufficient redundancy can compensate for the lack of the structured noise process.

\begin{table}[t]
    \centering
    \caption{The second occurrence of ``cat'' is corrupted to ``dog''.}
    \label{tab:corruption_4}

    \small
    \setlength{\tabcolsep}{6pt}
    \renewcommand{\arraystretch}{1.15}

    \begin{tabular}{
        @{}
        p{0.16\linewidth}
        p{0.66\linewidth}
        p{0.10\linewidth}
        @{}
    }
        \toprule

        \multicolumn{3}{@{}l@{}}{
            \textbf{Example 4: cat $\rightarrow$ dog}
            \hspace{1.2em}
            \textit{(2nd occurrence)}
        } \\[3pt]

        \midrule


        \multicolumn{3}{@{}p{\linewidth}@{}}{
            \textbf{Original}\quad
            Lily had a pretty cat.
            The \textbf{cat} was soft and white.
            Lily liked to pet her cat every day.
            One day the cat found a little mouse.
            Lily and the cat played in the garden.
        } \\[6pt]

        \multicolumn{3}{@{}p{\linewidth}@{}}{
            \textbf{Corrupted}\quad
            Lily had a pretty cat.
            The \textbf{dog} was soft and white.
            Lily liked to pet her cat every day.
            One day the cat found a little mouse.
            Lily and the cat played in the garden.
        } \\[3pt]

        \midrule


        \textbf{Model}
        &
        \textbf{Infilled text}
        &
        \textbf{Repair?}
        \\

        \midrule

        VSDD
        &
        \ldots The \textbf{cat} was soft and white \ldots

        \vspace{1pt}
        {\footnotesize\itshape
        Successful in 5/5 trials.}
        &
        Yes
        \\[4pt]

        D3PM-Uniform
        &
        \ldots The \textbf{a} was soft and white \ldots

        \vspace{1pt}
        {\footnotesize\itshape
        Occasionally generates the article ``a'' instead of ``cat''.}
        &
        No
        \\[4pt]

        MDLM
        &
        \ldots The \textbf{cat} was soft and white \ldots

        \vspace{1pt}
        {\footnotesize\itshape
        Successful in 5/5 trials.}
        &
        Yes
        \\

        \bottomrule
    \end{tabular}
\end{table}

\begin{table}[t]
    \centering
    \caption{Qualitative example for the train $\rightarrow$ ball corruption.
    The third occurrence of ``train'' is corrupted to ``ball'', and each
    model attempts to recover the original token.}
    \label{tab:corruption_5}

    \small
    \setlength{\tabcolsep}{6pt}
    \renewcommand{\arraystretch}{1.15}

    \begin{tabular}{
        @{}
        p{0.16\linewidth}
        p{0.66\linewidth}
        p{0.10\linewidth}
        @{}
    }
        \toprule

        \multicolumn{3}{@{}l@{}}{
            \textbf{Example 5: train $\rightarrow$ ball}
            \hspace{1.2em}
            \textit{(3rd occurrence)}
        } \\[3pt]

        \midrule


        \multicolumn{3}{@{}p{\linewidth}@{}}{
            \textbf{Original}\quad
            Ben had a toy train.
            Ben loved his toy train.
            Every day Ben played with the toy \textbf{train} in his room.
            One day Ben took the toy train to show his friend.
            His friend liked the toy train too.
        } \\[6pt]

        \multicolumn{3}{@{}p{\linewidth}@{}}{
            \textbf{Corrupted}\quad
            Ben had a toy train.
            Ben loved his toy train.
            Every day Ben played with the toy \textbf{ball} in his room.
            One day Ben took the toy train to show his friend.
            His friend liked the toy train too.
        } \\[3pt]

        \midrule


        \textbf{Model}
        &
        \textbf{Infilled text}
        &
        \textbf{Repair?}
        \\

        \midrule

        VSDD
        &
        \ldots the toy \textbf{train} in his room \ldots

        \vspace{1pt}
        {\footnotesize\itshape
        Successful in 5/5 trials.}
        &
        Yes
        \\[4pt]

        D3PM-Uniform
        &
        \ldots the toy \textbf{train} in his room \ldots

        \vspace{1pt}
        {\footnotesize\itshape
        Successful in 2/5 trials.}
        &
        Partial
        \\[4pt]

        MDLM
        &
        \ldots the toy \textbf{train} in his room \ldots

        \vspace{1pt}
        {\footnotesize\itshape
        Successful in 5/5 trials.}
        &
        Yes
        \\

        \bottomrule
    \end{tabular}
\end{table}

Overall, we see that using D3PM-Uniform with an unstructured noise process frequently generates plausible but incorrect tokens (e.g., the pronoun "She" instead of "Lily", or the article "a" instead of "cat"). 
In contrast, VSDD's learned kernel provides intermediate token-level affinity structure, yielding substantially better repair than the uniform baseline.

\paragraph{Experiment 2: Subsequence Removal.}
In our second experiment, a contiguous middle subsequence (approximately 35\%--55\% of the story length, corresponding to 8--9 tokens) is removed and replaced with [MASK] tokens. 
The removed subsequence typically contains a narrative bridge connecting the story's setup to its conclusion (see the example below). 
For all stories, we measure the \textit{token overlap} (i.e., the fraction of in-filled tokens that exactly match the original removed subsequence).
Table \ref{tab:span-infilling-results} shows the aggregated results.

\begin{table}[t]
    \centering
    \caption{Removed subsequences used for re-drafting.
    Each example masks a contiguous subsequence of tokens from the original story.}
    \label{tab:removed-spans}

    \setlength{\tabcolsep}{8pt}
    \renewcommand{\arraystretch}{1.15}

    \begin{tabular}{@{}l p{0.54\linewidth} c@{}}
        \toprule
        \textbf{Story} &
        \textbf{Removed subsequence} &
        \textbf{subsequence length} \\
        \midrule

        Lily \& red ball &
        \textit{``red ball. She loved to play with the''} &
        9 \\

        Tom \& Max &
        \textit{``in the park. Every morning Tom took''} &
        8 \\

        Sam \& blue car &
        \textit{``liked to play with his blue car. One''} &
        9 \\

        \bottomrule
    \end{tabular}
\end{table}

From Table \ref{tab:span-infilling-results} we see that across the three stories VSDD achieves the highest average overlap (24.3\%) and produces the most coherent bridges, followed by MDLM (18.4\%) and D3PM-Uniform (16.3\%).  
The low absolute values of the token overlaps are expected since there are many valid ways to bridge the narrative gap (so the metric provides a lower bound on generation quality). 

In Table \ref{tbl:bridge}, we provide an example from this experiment.
From the example, we can see that VSDD preserves both the key entity ("red ball") and the narrative structure ("she loved ... She took her"), yielding the highest token overlap. 
D3PM-Uniform produces incoherent repetition ("The red liked red new toy"). 
MDLM generates a grammatical but nonsensical phrase ("ball with many hair").
Qualitatively, VSDD more consistently preserves key entities from the surrounding context and generates grammatical continuations.

\begin{table}[t]
    \centering
    \caption{Span infilling performance.
    Token Overlap measures agreement between the generated and original
    removed spans.}
    \label{tab:span-infilling-results}

    \setlength{\tabcolsep}{10pt}
    \renewcommand{\arraystretch}{1.12}

    \begin{tabular}{@{}lcc@{}}
        \toprule
        \textbf{Model} &
        \textbf{Token Overlap} $\uparrow$ \\
        \midrule

        VSDD  & $\mathbf{0.243 \pm 0.127}$  \\
        D3PM-Uniform & $0.163 \pm 0.159$           \\
        MDLM         & $0.184 \pm 0.082$           \\
        \bottomrule
    \end{tabular}
\end{table}

\begin{table}[t]
\caption{Example: Subsequence Removal.}\label{tbl:bridge}
    \centering
    \small
    \setlength{\tabcolsep}{6pt}
    \renewcommand{\arraystretch}{1.15}

    \begin{tabular}{
        @{}
        p{0.17\linewidth}
        p{0.64\linewidth}
        p{0.11\linewidth}
        @{}
    }
        \toprule

        \multicolumn{3}{@{}l@{}}{
            \textbf{Example: Lily story}
            \hspace{1.2em}
            \textit{(9 tokens removed)}
        } \\[3pt]
        \midrule

        \multicolumn{3}{@{}p{\linewidth}@{}}{
            \textbf{Original}\quad
            \ldots Lily. Lily had a big
            \textbf{red ball. She loved to play with the}
            red ball every day \ldots
        } \\[6pt]

        \multicolumn{3}{@{}p{\linewidth}@{}}{
            \textbf{Removed span}\quad
            \textit{``red ball. She loved to play with the''}
        } \\[3pt]
        \midrule

        \textbf{Model}
        &
        \textbf{Infilled span}
        &
        \textbf{Overlap} $\uparrow$
        \\
        \midrule

        VSDD
        &
        ``\textbf{red ball that she loved. She took her}''
        &
        $\mathbf{42.2\%}$
        \\[4pt]

        D3PM-Uniform
        &
        ``\textbf{red toy. The red liked red new toy}''
        &
        $11.1\%$
        \\[4pt]

        MDLM
        &
        ``\textbf{ball with many hair. She played with the}''
        &
        $8.9\%$
        \\

        \bottomrule
    \end{tabular}
\end{table}

\clearpage
\subsection{Playlist Generation}\label{app:redrafting_playlist}

Unlike in the molecular domain discussed in Section \ref{sec:results}, analysing the re-drafting capabilities of VSDD by visually inspecting the learned forward noise matrices in the playlist generation domain is not feasible.
This is because the different token substitutions are not based on the underlying world knowledge  but on the subjective (unobservable) user preferences. 
In this section, we show that the model can successfully repair local rule-based corruptions that typically reduce playlist quality.

To evaluate whether the learned forward process captures meaningful structural properties of playlist sequences, we design a targeted repair experiment. 
To this end, we introduce controlled corruptions into held-out playlists, and we measure the ability of the trained model to restore the structural coherence of the playlists via conditional reverse diffusion.

In particular, we define three corruption operators, each modifying exactly one track (i.e., three semantic tokens) per sequence:
\begin{itemize}[leftmargin=0.99cm] 
    \item \textbf{Track duplication}. A source track position $s$ is selected uniformly at random, and a distinct destination position $d \neq s$ is selected uniformly at random. The SID triplet corresponding to the track at position $d$ is overwritten with a copy of the triplet of the track at position $s$, simulating a repeated-track artifact. 

    \item \textbf{Transition disruption}. We identify the consecutive track pair $(i, i{+}1)$ with the highest audio-feature\footnote{In our dataset, audio features are 16-dimensional vectors of normalized acoustic descriptors (energy, danceability, valence, tempo, etc.) aggregated per SID triplet.} cosine similarity (i.e., the smoothest transition) across all adjacent pairs in the playlist. We then scan all remaining tracks $j \notin \{i, i{+}1\}$ and select the one whose audio-feature vector has the lowest cosine similarity to track $i$. The SID triplet corresponding to the track at position $i{+}1$ is then overwritten with the triplet of this maximally dissimilar track, creating a jarring transition at the previously smoothest point, thereby reducing consumption \citep{Moor_CIKM_2023}.
    The corrupted region is $\{3(i{+}1), 3(i{+}1){+}1, 3(i{+}1){+}2\}$. 

    \item \textbf{Energy misplacement}.  We identify the track with the highest energy value across all positions and swap it with the track at the very first position of the playlist (the playlist opening). 
    If the highest-energy track is already at the first position, the lowest-energy track is swapped to this position instead. 
    The corrupted region is $\{0, 1, 2\}$ (the triplet corresponding to the opening track only). 
    Note that while the swap modifies two track positions, only the track in the first position is designated for repair, testing whether the model can restore an appropriate opening track given one-sided (right-only) context.
\end{itemize}

For each corrupted sequence, we construct a binary generation mask $\mathbf{m} \in \{0,1\}^L$ with $m_\ell = 1$ at the three corrupted token positions and $m_\ell = 0$ elsewhere. The corrupted sequence $\tilde{\mathbf{x}}$ and mask $\mathbf{m}$ are passed to the conditional reverse diffusion procedure: masked positions are initialized with tokens drawn uniformly from the valid SID vocabulary, while unmasked positions retain the corrupted values throughout. The model then performs $T = 50$ reverse diffusion steps using the learned transition matrices from the Stackelberg-trained forward process. At each step, reverse sampling probabilities are computed from the model's predicted logits and the learned posterior, but token updates are applied only at the three masked positions. The surrounding 147 tokens serve as fixed conditioning context.

We report three quantities, each computed over $200$ corrupted validation playlists:
\begin{itemize}
    \item \textbf{Repair rate}. The fraction of corrupted token positions at which the model's output differs from the corrupted value. 
    A high repair rate indicates that the model recognizes the corruption as inconsistent with the surrounding context.
    
    \item \textbf{Audio smoothness}. The mean cosine similarity between the 16-dimensional audio-feature vectors of the consecutive tracks, computed over all adjacent pairs in the playlist. 
    We report this for the original (O), corrupted (C), and repaired (R) sequences.

    \item \textbf{Style diversity}. The number of unique genre/style descriptor tags across all tracks in the sequence, obtained from per-track metadata (with approximately 50 weighted descriptors per track from the editorial taxonomy). 
    Reported for O, C, and R.
\end{itemize}

Table~\ref{tab:corruption_repair} summarizes the repair outcomes.

\begin{table}[t]
    \centering
    \caption{Repair results for different corruption types.}
    \label{tab:corruption_repair}
    \setlength{\tabcolsep}{4pt}
    \begin{tabular}{@{}lrrrrrrr@{}}
        \toprule
        & & \multicolumn{3}{c}{Smooth}
          & \multicolumn{3}{c}{Diversity} \\
        \cmidrule(lr){3-5}
        \cmidrule(lr){6-8}
        Corruption & Repair rate
        & O & C & R
        & O & C & R \\
        \midrule
        Duplication
        & 78.9\%
        & 0.9800 & 0.9802 & 0.9807
        & 2477 & 2453 & 2476 \\
        Transition disruption
        & 86.1\%
        & 0.9800 &  0.9767 & 0.9792
        & 2510 & 2497 & 2520 \\
        Energy misplacement
        & 85.3\%
        & 0.9800 & 0.9797 & 0.9799
        & 2477 & 2477 & 2481 \\
        \bottomrule
    \end{tabular}
\end{table}

First, we see that VSDD actively repairs corrupted positions in 78--86\% of cases, with the highest repair rate for transition disruption (86.1\%), where the corrupted track is maximally inconsistent with its immediate neighbours. 
This suggests the model has learned local coherence constraints: It detects that a track with very different audio characteristics does not belong between its neighbours and proposes a more suitable replacement.

For transition disruption, the corruption reduces audio smoothness from 0.9800 to  0.9767. After repair, the smoothness recovers to 0.9792. 
The model does not fully recover the original smoothness because it is not constrained to reproduce the original track. 
Instead, it generates any track consistent with the learned distribution conditioned on the context.

Track duplication yields a slightly lower repair rate (78.9\%), consistent with the observation that a duplicated track may be contextually plausible at its destination: Unlike transition disruption, duplication does not necessarily create a local inconsistency, especially when the track is highly relevant for the user.

Energy misplacement achieves an intermediate repair rate (85.3\%) despite having access to only right-side context at the very first position of the playlist. 
The model must infer an appropriate opening track from the subsequent sequence alone, which is a more complex task as users typically interact with the playlist from left to right (so the prefix context is often more informative when making the track allocation decision than the suffix-context).

Audio smoothness and style diversity show minimal variation across conditions because each corruption modifies only one of approximately 49 tracks. The smoothness metric averages over 48 consecutive pairs, diluting the local effect of a single corrupted transition. Style diversity is similarly insensitive: each track contributes dozens of tags, and the union over 49 tracks saturates at approximately 2,500 unique descriptors regardless of single-track substitutions.

The repair experiment provides evidence that the Stackelberg-trained diffusion model captures local sequential structure in playlist data. 
The learned forward process, optimized to improve the denoiser's generalization, produces a reverse model capable of context-sensitive infilling. 
The high repair rate for transition disruption (86.1\%) demonstrates that the model encodes audio-feature continuity as an implicit constraint, even though audio features are not explicitly provided during training as the model operates solely on discrete SID tokens derived from track embeddings via locality-sensitive hashing. 
The correlation between embedding cosine similarity and learned transition probabilities (Spearman $\rho = 0.74$ for the learned noise vs. embedding similarity, see Figure \ref{fig:result-c}) provides a mechanistic explanation: The forward process preferentially corrupts tokens toward embedding-similar alternatives, and the reverse process inherits this inductive bias.

\section{Degenerate Noise}\label{app:degenerate_noise}
Table \ref{tab:kernel-collapse} illustrates the per-row entropy and the dominant transitions of $Q_{1,\phi}$ at $t{=}1$ of the most frequent tokens learned by naive joint optimisation of $\phi$ and $\theta$ (D3PM-Reinforce, see Section \ref{sec:results}).
The uniform maximum entropy in this example is $\ln(V{-}1) \approx 7.62$.
From the table we can see that the top rows exhibit near-deterministic transitions, indicating degenerate noise process collapse targeting the high-frequency tokens.

\begin{table}[h]
    \centering
    \caption{Naive joint optimisation of $\phi$ and $\theta$ learns trivial near-deterministic noise processes.}
    \label{tab:kernel-collapse}
    \begin{tabular}{lccc}
        \toprule
        \textbf{Token} & \textbf{Row Entropy}
        & \textbf{Top Target} & $\boldsymbol{P(i\rightarrow j)}$ \\
        \midrule
        \texttt{.}   & 0.023 & \texttt{to}   & 0.998 \\
        \texttt{the} & 0.062 & \texttt{t}    & 0.996 \\
        \texttt{and} & 0.095 & \texttt{it}   & 0.993 \\
        \texttt{,}   & 0.096 & \texttt{that} & 0.993 \\
        \texttt{to}  & 0.111 & \texttt{.}    & 0.992 \\
        \texttt{it}  & 0.115 & \texttt{was}  & 0.991 \\
        \texttt{in}  & 0.216 & \texttt{,}    & 0.983 \\
        \texttt{was} & 0.232 & \texttt{and}  & 0.981 \\
        \texttt{t}   & 0.439 & \texttt{was}  & 0.963 \\
        \texttt{she} & 0.552 & \texttt{that} & 0.952 \\
        \texttt{She} & 0.730 & \texttt{They} & 0.935 \\
        \texttt{so}  & 0.863 & \texttt{in}   & 0.921 \\
        \texttt{out} & 1.733 & \texttt{up}   & 0.829 \\
        \texttt{on}  & 1.944 & \texttt{for}  & 0.806 \\
        \bottomrule
    \end{tabular}
\end{table}

\end{document}